\documentclass[sigconf]{acmart}

\AtBeginDocument{%
  }

\usepackage{multirow}
\usepackage{enumitem}

\acmConference[WSDM '27]
{The 20th ACM International Conference on Web Search and Data Mining}
{February 15--19, 2027}
{Hong Kong}

\begin{document}

\title{PailitaoGR: Latent Think-with-Images for Generative Image Retrieval}

% The anonymous option hides author information during review.
% \author{Anonymous Author(s)}

\author{Xiaomeng Fan}
\authornote{Equal Contribution.}
\affiliation{
  \institution{Alibaba Group}
  \city{Hangzhou}
  \country{China}}
\email{fanxiaomeng.fxm@taobao.com}

\author{Yueran Liu}
\authornotemark[1]
\authornote{Project Leader.}
\affiliation{
  \institution{Alibaba Group}
  \city{Hangzhou}
  \country{China}}
\email{tianer.lyr@taobao.com}

\author{Shengyu Zhou}
% \authornotemark[1]
\affiliation{
  \institution{Alibaba Group}
  \city{Hangzhou}
  \country{China}}
\email{zhoushengyu.zsy@taobao.com}

\author{Chenghan Fu}
% \authornotemark[1]
% \authornote{Project Leader.}
\affiliation{
  \institution{Alibaba Group}
  \city{Hangzhou}
  \country{China}}
\email{fuchenghan.fch@taobao.com}

\author{Wanxian Guan}
\affiliation{
  \institution{Alibaba Group}
  \city{Hangzhou}
  \country{China}}
\email{wanxian.gwx@taobao.com}

\author{Feng Li}
\affiliation{
  \institution{Alibaba Group}
  \city{Hangzhou}
  \country{China}}
\email{adam.lf@taobao.com}

\author{Chuan Yu}
\affiliation{
  \institution{Alibaba Group}
  \city{Hangzhou}
  \country{China}}
\email{yuchuan.yc@taobao.com}

\author{Jian Xu}
\affiliation{
  \institution{Alibaba Group}
  \city{Hangzhou}
  \country{China}}
\email{xiyu.xj@taobao.com}

\author{Bo Zheng}
\authornote{Corresponding Author.}
\affiliation{
  \institution{Alibaba Group}
  \city{Hangzhou}
  \country{China}}
\email{bozheng@alibaba-inc.com}

\begin{abstract}
 Generative retrieval has demonstrated strong performance by directly generating product semantic identifiers (SIDs). 
 Extending this paradigm to image search, however, is nontrivial because real-world query images contain diverse information, including the search target, useful auxiliary evidence, and irrelevant visual content.
 This requires the model to identify and focus on the search target while selectively utilizing auxiliary evidence. 
In this paper, we propose \textbf{PailitaoGR}, a \emph{Latent Think-with-Images} method for generative image retrieval, which internalizes target-focused perception and selective auxiliary-evidence utilization into a the generative retrieval model, enabling \textit{Zooming without Cropping} and \textit{Reading without OCR}.
Specifically, we design a target-focused perception mechanism that identifies and enhances visual tokens of the search target, consisting of a target Enhancer and a learning strategy based on on-policy distillation and attention guidance loss, enabling the model to focus on search-target regions.
We also design a selective auxiliary-evidence utilization mechanism that identifies and enhances visual tokens of auxiliary evidence, including an auxiliary enhancer and an in-capacity incremental contrastive distillation strategy, enabling the model to exploit auxiliary evidence.
We construct training and validation sets sampled from real-world online image-search logs.
Experiments show that our method outperforms existing baselines by an average of 13.8\%, validating its effectiveness.
% Extending this paradigm to image search, however, is nontrivial because real-world query images contain heterogeneous visual information, and demand fine-grained same-product matching. 
\end{abstract}

\vspace{-5em}

% Add the ACM Computing Classification System concepts generated at
% https://dl.acm.org/ccs before the camera-ready submission.
% \begin{CCSXML}
% ...
% \end{CCSXML}
% \ccsdesc[500]{Information systems~Web searching and information discovery}

\maketitle

\section{Introduction}

Large-scale e-commerce image search, such as Pailitao, Taobao's image search system, allows users to retrieve target items directly from real-world query images~\cite{chen2026pailitao}.
In this work, we explore generative image retrieval for this task, motivated by the success of generative retrieval~\cite{chen2026onesearchv2} and generative recommendation~\cite{deng2025onerec,han2025mtgr}.
Real-world query images often contain heterogeneous visual information, including the search target, useful auxiliary cues such as brand and model information, and irrelevant content such as watermarks and other objects.
Accordingly, effective generative image retrieval requires two complementary capabilities: identifying and focusing on the search target, and selectively utilizing relevant auxiliary evidence for fine-grained item identification.

Recent \emph{Think-with-Images} paradigms address complex visual understanding by interacting with images through tools, such as visual grounding, cropping, and OCR~\cite{zheng2026deepeyes,zhang2025chain}.
Directly adopting this paradigm for online generative image retrieval, however, faces two challenges.
First, explicit tool invocation introduces additional computation and multi-step inference, which conflicts with the strict latency requirements of online retrieval.
Second, auxiliary information obtained from these tools may be irrelevant or misleading, while some useful information may exceed the model's perceptual capacity due to limited image resolution and model size, making indiscriminate capability transfer potentially harmful.

\begin{figure*}[t]
    \centering
    \includegraphics[width=0.85\linewidth]{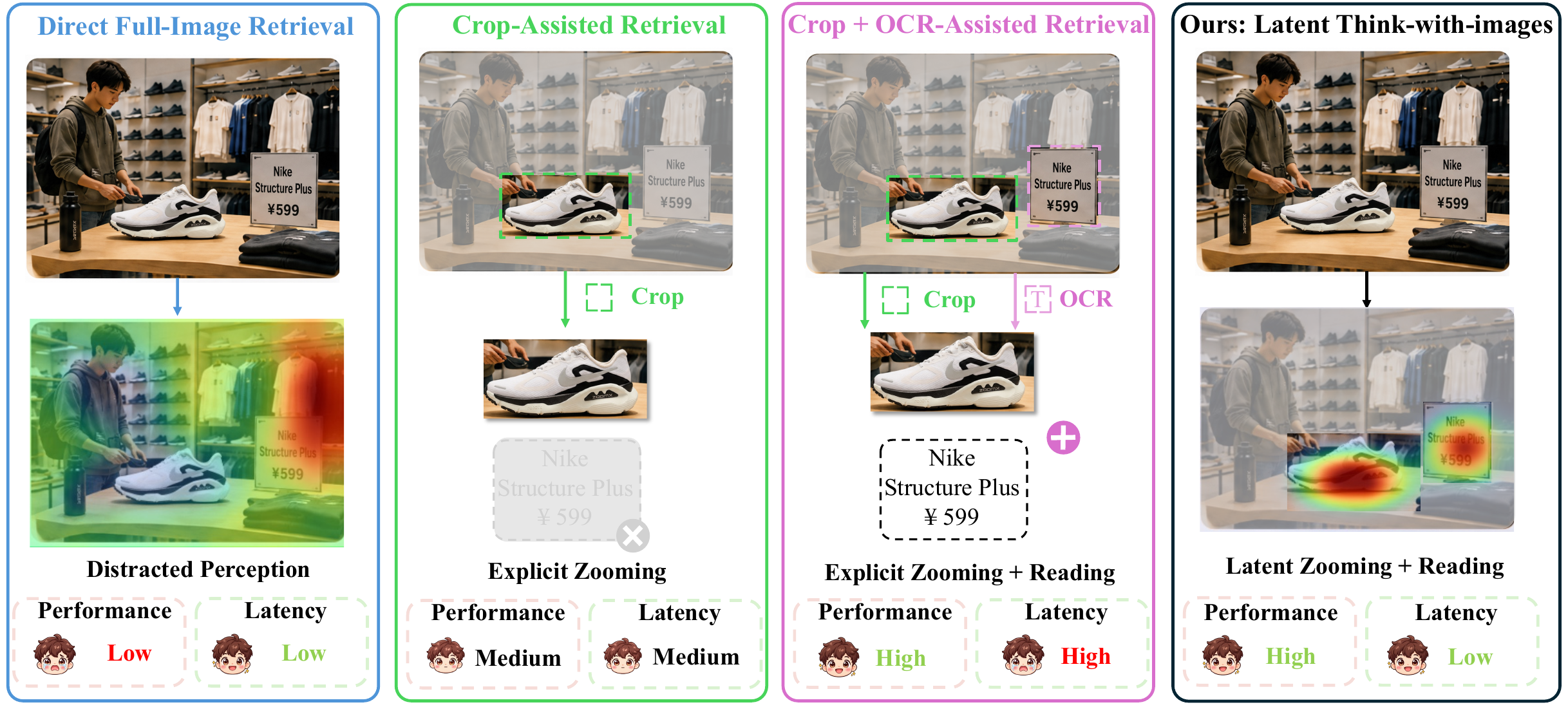}
    % \vspace{-2em}
    \caption{Comparison of direct, tool-assisted, and latent Think-with-Images paradigms for generative image retrieval. Our approach internalizes tool-enabled visual capabilities without extra inference latency.}
    \label{fig:figure1}
\end{figure*}

In this paper, we propose \textbf{PailitaoGR}, a \emph{Latent Think-with-Images} method for generative image retrieval, which internalizes target focusing and auxiliary-evidence utilization into a model that operates solely on the original query image, as illustrated in Fig.~\ref{fig:figure1}.
Specifically, PailitaoGR contains two capability internalization mechanisms: a target-focused perception mechanism to achieve \textit{Zooming without Cropping}, and a selective auxiliary-evidence utilization mechanism to achieve \textit{Reading without OCR}.
% In target-focused perception mechanism, we introduce an target enhancer that identifies and enhances visual tokens of the search target in the query image.
In the target-focused perception mechanism, we introduce a target enhancer that identifies and enhances visual tokens of the search target in the query image.
We further construct a Crop Teacher that takes only the cropped search-target region as input, providing a cleaner and more target-focused prediction.
Through on-policy distillation, our model, which takes the original full query image as input, implicitly learns this target-focused prediction behavior.
Beyond implicitly learning to focus, we introduce region-of-target (ROT)-based and entropy-based attention guidance to explicitly encourage SID generation to remain centered on the search-target region throughout decoding.
The ROT loss directs visual attention toward the search-target region, and the entropy loss encourages broader attention for coarse predictions and progressively more concentrated attention for fine-grained discrimination.

% In selective auxiliary-evidence utilization mechanism, particularly OCR-derived textual cues, we introduce an auxiliary enhancer that identifies and enhances useful textual cues relevant to the search target. 
In the selective auxiliary-evidence utilization mechanism, we introduce an auxiliary enhancer that identifies and enhances useful auxiliary evidence, particularly visual cues containing OCR-derived textual information relevant to the search target.
We further construct an OCR Teacher that takes the cropped search-target region and pre-extracted OCR text as input.
To selectively transfer this capability, we develop an in-capacity incremental contrastive distillation strategy, which activates OCR supervision only when the textual information benefits target SID prediction and is accessible to the current model, and transfers only the capability gain of the OCR Teacher over the Crop Teacher.
In this way, the model learns to exploit useful textual evidence while avoiding irrelevant, conflicting, or inaccessible information.

We construct both the training and validation sets from online image-search logs on Pailitao. Compared with existing generative image retrieval and contrastive retrieval methods, our method achieves the best performance, demonstrating its effectiveness. We further compare with a Crop Teacher trained and evaluated on target crops, and a stronger OCR Teacher that additionally incorporates pre-extracted textual information. Our model consistently outperforms both teachers, indicating that target-focusing and auxiliary-evidence utilization capabilities are effectively internalized into the model. Upon acceptance, we will release the curated datasets and trained models.

% \vspace{-em}
\begin{itemize}[leftmargin=*]

    \item  We propose \textbf{PailitaoGR}, a \emph{Latent Think-with-Images} method for generative image retrieval, which internalizes target-focused perception and selective auxiliary-evidence utilization into a model, enabling it to focus on the search target and effectively utilize useful auxiliary cues without additional tool calls.

    \item We design a target-focused perception mechanism, which identifies and enhances visual tokens of the search target through a target enhancer and an attention-guided on-policy distillation strategy, effectively internalizing target-focused perception.

\item We develop a selective auxiliary-evidence utilization mechanism, which identifies and enhances useful auxiliary evidence through an auxiliary enhancer and an in-capacity incremental contrastive distillation strategy, selectively internalizing auxiliary-evidence utilization.

   % \item We develop an target enhancer together with on-policy Crop Teacher distillation and attention guidance loss, which effectively internalizes target-focused perception.

   %  \item We introduce an auxiliary enhancer  and an in-capacity incremental contrastive distillation strategy, which selectively internalizes auxiliary-evidence utilization.

    \item We construct training and validation sets covering seven representative categories from real-world online image-search logs. Upon acceptance, we will release the datasets and trained models to support further research on generative image retrieval.
    
\end{itemize}

% First, although cropping and OCR can provide target localization and text recognition, online retrieval cannot afford additional tool calls or multi-step inference, requiring these abilities to be internalized into a single full-image forward pass for “zooming without cropping” and “reading without OCR.” Second, auxiliary cues may be irrelevant, conflict with product appearance, or exceed the model’s perceptual capacity due to limited resolution and model size, making indiscriminate knowledge transfer potentially harmful.

% Generative image retrieval requires the model to first infer the user’s intent and locate the focal product, and then selectively extract relevant evidence from the image to capture the user’s preferences.

% Generative retrieval encodes candidate products into discrete identifiers and directly generates the SID corresponding to a query, showing strong potential for large-scale product search. Extending this paradigm to image queries enables users to retrieve identical products from real-world photos. Unlike concise text queries, such images often mix the target product with useful cues, such as brand and model information, and irrelevant content, such as background text, watermarks, and other objects. Accordingly, generative image retrieval requires the model to first identify the user-focused product and then selectively exploit auxiliary evidence for fine-grained discrimination, while suppressing irrelevant or conflicting information.

\section{Related Works}

\subsection{Representation-based Vision Search}

Representation-based vision search encodes query images and products into dense representations and retrieves relevant products through nearest-neighbor search. Traditional e-commerce representation learning relies on dual-flow architectures for cross-modal alignment between visual and textual product content~\cite{chia2022contrastive,dai2024uniembedding,liang2025uniecs}
% \cite{chia2022contrastive,yu2022commercemm,chia2022fashionclip,dai2024uniembedding,liang2025uniecs}.
% More recently, MLLM-based embedding models adapt pretrained MLLMs to produce dense representations from image, text, or interleaved image--text inputs, such as E5-V, VLM2Vec, MM-Embed, GME, Qwen3-VL-Embedding, and the MOON series
Recent studies leverage pretrained multimodal large language models for representation learning, 
% adapting their multimodal understanding capabilities to produce embeddings from arbitrary combinations of images and text, 
including  VLM2Vec, MM-Embed, GME, Qwen3-VL-Embedding, and the MOON series~\cite{jiang2024vlm2vec,lin2024mmembed,zhang2024gme,li2026qwen3vlembedding,zhang2025moon,nie2025moon2}. 
The emerging Think-Then-Embed paradigm further exploits the generative and reasoning capabilities of MLLMs to derive intermediate semantic contexts before embedding extraction~\cite{yan2025o1embedder,cui2025thinkthenembed,hao2026trace,jiang2026embedrl,wu2026moon3}.

% \vspace{-1em}
\subsection{Generative Retrieval}

Generative retrieval reformulates retrieval as an autoregressive generation problem.
Generative retrieval first encodes each item into a semantic identifier (SID), commonly constructed using clustering- or quantization-based methods such as RQ-VAE and FSQ~\cite{lee2022rqvae,mentzer2023fsq}. A generative model is then trained to directly generate the SID of relevant items conditioned on the input query.

Recent text-based generative search studies mainly explore three directions. First, several methods improve semantic alignment between queries, products, and SIDs through better identifier construction and contrastive constraints, including GenR-PO, GRAM, and CQ-SID~\cite{li2024genrpo,pang2025gram,zhu2026cqgr}. Second, recent works enhance query understanding by modeling explicit or latent user intent, such as context-aware reasoning, self-distilled thought augmentation, and category-guided latent intent learning~\cite{liu2025contextgr,chen2026onesearchv2,zhang2026calir}. Third, preference-aware methods further align SID generation with user behaviors and downstream search objectives through preference optimization, reinforcement learning, and value-aware ranking~\cite{chen2025onesearch,chen2026raddpo,zhan2026tsgr}.

Beyond textual queries, generative retrieval has also been explored for image and multimodal search.
IRGen introduces a generative image retrieval framework, while GENIUS further extends generative retrieval to multimodal queries; both demonstrate their effectiveness on public benchmarks~\cite{zhang2024irgen,kim2025genius}.
OneVision develops the first industrial generative retrieval framework for e-commerce vision search, where the generative retriever takes cropped query images as input~\cite{zheng2025onevision}.
In contrast, to our knowledge, we present the first industrial generative retrieval framework that operates directly on original query images, enabling effective target focusing and auxiliary-evidence utilization through \emph{Latent Think-with-Images}.

\begin{figure*}[t]
    \centering
    \includegraphics[width=0.85\linewidth]{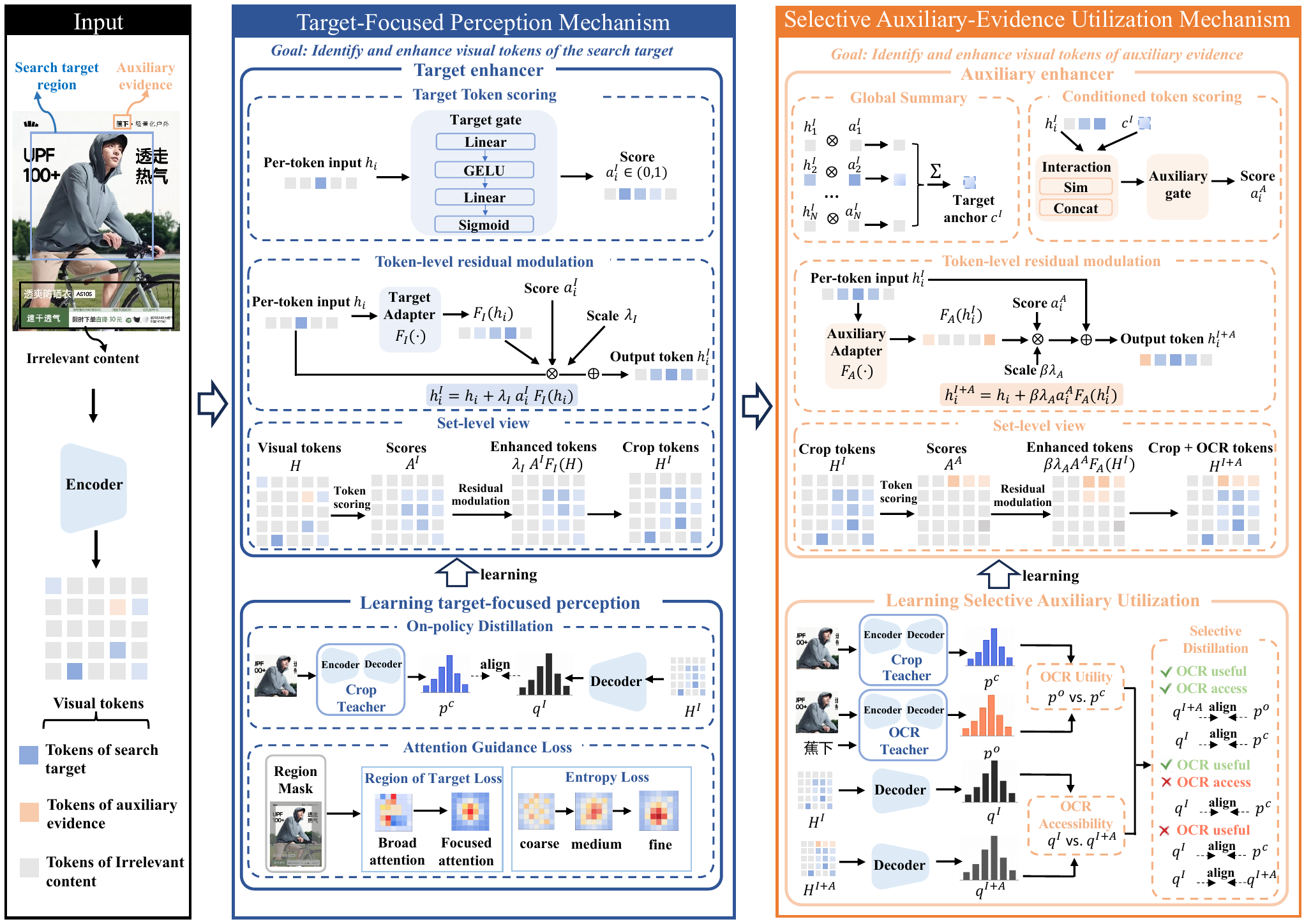}
    \caption{Framework of our method.}
    \label{fig:figures2}
\end{figure*}

% \begin{figure*}[t]
%     \centering
%         \includegraphics[width=\linewidth]{figures/Figuer2.pdf}
%     \caption{Framework of our method.}
%     \label{fig:figures2}
% \end{figure*}

\section{Method}
\label{sec:method}

We propose \textbf{PailitaoGR}, a \emph{Latent Think-with-Images} method for generative image retrieval.
We first formulate the generative image retrieval task and present an overview of the proposed method.
We then introduce target-focused perception mechanism  and selective auxiliary-evidence utilization mechanism, followed by the inference.

% \vspace{-1em}
\subsection{Formulation}
\label{sec:formulation}

Let $\mathcal{D}=\{d_j\}_{j=1}^{M}$ denote a product corpus, where each product $d_j$ is assigned a semantic identifier (SID)
\begin{equation}
\mathbf{s}_j=(s_{j,1},s_{j,2},\ldots,s_{j,L}),
\end{equation}
consisting of a sequence of discrete semantic tokens.
Given a query image $\mathbf{x}$, generative image retrieval directly models the conditional generation probability of a candidate SID as
\begin{equation}
p_{\theta}(\mathbf{s}\mid\mathbf{x})
=
\prod_{t=1}^{L}
p_{\theta}
\left(
s_t
\mid
\mathbf{x},
\mathbf{s}_{<t}
\right),
\label{eq:generative_retrieval}
\end{equation}
where $\mathbf{s}_{<t}=(s_1,\ldots,s_{t-1})$ denotes the previously generated SID tokens.

At inference, the generative model autoregressively generates a set of high-probability candidate SIDs, which is modeled as
\begin{equation}
\hat{\mathcal{S}}_{K}
=
\operatorname{TopK}_{\mathbf{s}\in\mathcal{S}}
p_{\theta}(\mathbf{s}\mid\mathbf{x}),
\label{eq:topk_sid}
\end{equation}
where $\hat{\mathcal{S}}_{K}$ contains the top-$K$ generated SIDs, which are subsequently mapped to their corresponding products for retrieval.

Real-world query images typically contain heterogeneous visual information, including the search target, potentially useful auxiliary evidence, and irrelevant content.
Consequently, effective generative image retrieval requires the model to both identify and focus on the search target and selectively utilize complementary evidence for fine-grained item identification.

To this end, we propose \textbf{PailitaoGR}, a \emph{Latent Think-with-Images} method that internalizes these two capabilities into a generative retriever operating solely on the original query image.
% As illustrated in Fig.~\ref{fig:figures2}, PailitaoGR realizes them through two core mechanisms: \textbf{Target-Focused Perception} to enable the model  to focus on the search target in complex query images and \textbf{Selective Auxiliary-Evidence Utilization}.
As illustrated in Fig.~\ref{fig:figures2}, PailitaoGR realizes them through two core mechanisms: \textbf{Target-Focused Perception} to enable the model to focus on the search target in complex query images and \textbf{Selective Auxiliary-Evidence Utilization} to selectively exploit  auxiliary evidence.

% \vspace{-1em}
\subsection{Target-Focused Perception Mechanism}
\label{sec:item_internalization}

% Real-world query images usually contain the target product together with background objects and other irrelevant visual content.
% Directly decoding SIDs from such heterogeneous information may cause the model to allocate its limited modeling capacity to regions that are unrelated to the target product.
% To address this issue,
% To enable the model to focus on the target product in complex query images, we introduce \textbf{target-focused perception mechanism}.
We design an \textbf{Target Router} to identify and enhance  target-related visual tokens, and further introduce an \textbf{Target-Focused Perception Objective} to guide the model toward target-centric prediction and visual attention during SID generation.

% We introduce \textbf{target-focused perception mechanism}, which adaptively strengthens visual tokens associated with the target product while preserving the original full-image representation.
% Specifically, we first identify product-related tokens through an target enhancer and then inject task-specific feature updates through gated residual modulation.
% We further design prediction-level and attention-level objectives to encourage the model to rely primarily on the target product during SID generation.

\subsubsection{Target Enhancer}
The \textbf{Target Enhancer} consists of \textbf{Target Token Scoring}, which selects visual tokens of search target, and \textbf{Token-level Residual Modulation}, which enhances the selected visual tokens.
% \paragraph{Target Token scoring and residual modulation.}

Given a query image $\mathbf{x}$, the visual encoder produces a set of visual tokens
\begin{equation}
H
=
E_{\theta}(\mathbf{x})
=
\{h_i\}_{i=1}^{N},
\qquad
h_i\in\mathbb{R}^{d}.
\label{eq:item_visual_tokens}
\end{equation}
We employ a lightweight \textbf{Target Token Scoring} to independently estimate the relevance of each visual token to the search target, 
\begin{align}
a_i^{I}
&=
\sigma(w_I^{\top}
\operatorname{GELU}
\left(
W_I\operatorname{LN}(h_i)
\right)),
\label{eq:item_weight}
\end{align}
where $W_I$ and $w_I$ are learnable parameters, $\operatorname{LN}(\cdot)$ is the layer normalization, and $\sigma(\cdot)$ denotes the sigmoid function.
$a_i^{I}$ measures the relevance of visual token $i$ to the search target.
% $a_i^{I}$ indicates how strongly token $i$ should be adapted toward product identification.

% We adopt sigmoid rather than softmax because the target product may occupy an arbitrary number of visual tokens.
% Instead of forcing different tokens to compete for a fixed probability mass, sigmoid allows all product-related tokens to receive high relevance scores simultaneously.
% Therefore, 

% Simply rescaling $h_i$ only changes its magnitude, providing limited ability to enhance the visual representation itself.
Simply rescaling $h_i$ according to $a_i^{I}$ only changes its magnitude without altering its feature direction, and its effect can be further weakened by subsequent normalization.
We introduce \textbf{Token-level Residual Modulation} to enhance the selected visual tokens through learnable residual feature updates by
% We therefore introduce \textbf{Token-level Residual Modulation}, which learns a residual feature transformation for each visual token:
\begin{equation}
F_I(h_i)
=
W_{I,2}
\operatorname{GELU}
\left(
W_{I,1}\operatorname{LN}(h_i)
\right).
\label{eq:item_residual_transform}
\end{equation}
The selected visual tokens are then enhanced through
\begin{equation}
h_i^{I}
=
h_i
+
\lambda_I
a_i^{I}
F_I(h_i),
\label{eq:item_residual_modulation}
\end{equation}
where $\lambda_I$ controls the magnitude of the residual enhancement.
Accordingly, the crop tokens that are target-enhanced visual representation  are computed as
\begin{equation}
H^{I}
=
\{h_i^{I}\}_{i=1}^{N}.
\label{eq:item_enhanced_tokens}
\end{equation}
In this way, token-level residual modulation enhances the selected visual tokens of search target while preserving their original visual information through the residual connection.

\subsubsection{Target-Focused Perception Objective.}

The most straightforward way to optimize the target enhancer is to use the ground-truth SID as supervision.
We denote the  SID distribution predicted from $H^{I}$ as
\begin{equation}
q_t^{I}(\cdot)
=
p_{\theta}
\left(
\cdot
\mid
H^{I},
\mathbf{s}_{<t}
\right).
\label{eq:item_student_distribution}
\end{equation}
The standard cross-entropy objective is
\begin{equation}
\mathcal{L}_{\mathrm{CE}}
=
-
\frac{1}{L}
\sum_{t=1}^{L}
\log
q_t^{I}
\left(
s_t^{*}
\mid
\mathbf{s}_{<t}^{*}
\right).
\label{eq:item_ce_loss}
\end{equation}
SID supervision only constrains the final prediction and cannot directly guide the model to focus on the search target in complex query images.
We introduce two objectives:
an \textbf{On-Policy Distillation Loss}, which implicitly transfers target-focused prediction behavior from a Crop Teacher, and a \textbf{Granularity-Aware Attention Guidance Loss}, which explicitly guides visual attention toward the search target during SID generation.

\paragraph{On-policy distillation.}
We train a \textbf{Crop Teacher} $p_{\phi_C}(\cdot)$ that receives the cropped image $\mathbf{x}^{C}$ of the search target.
Since most background regions are removed, the Crop Teacher provides a  reference of search target for SID prediction.
Conventional autoregressive distillation evaluates teacher and student under ground-truth prefixes, whereas inference conditions on the student's own predictions. At inference, once an SID token is predicted incorrectly, the model must continue generation from an erroneous prefix that is not encountered during distillation, making subsequent predictions less reliable and leading to a train--inference mismatch.

To reduce this discrepancy, we adopt \textbf{on-policy distillation}.
Given  $H^{I}$, we first generate a SID trajectory using the current student model that takes full image as input,
\begin{equation}
\hat{\mathbf{s}}
\sim
p_{\theta}
\left(
\mathbf{s}
\mid
H^{I}
\right).
\label{eq:item_rollout}
\end{equation}
At each decoding step $t$, both the Crop Teacher and the student are then queried using the same student-generated prefix $\hat{\mathbf{s}}_{<t}$:
\begin{align}
p_t^{C}(\cdot)
&=
p_{\phi_C}
\left(
\cdot
\mid
\mathbf{x}^{C},
\hat{\mathbf{s}}_{<t}
\right),
\label{eq:crop_teacher_distribution}
\\
q_t^{I}(\cdot)
&=
p_{\theta}
\left(
\cdot
\mid
H^{I},
\hat{\mathbf{s}}_{<t}
\right).
\label{eq:item_onpolicy_distribution}
\end{align}

In practice, we distill the top-$K$ candidates predicted by the Crop Teacher.
Let $\mathcal{K}_t$ denote the corresponding candidate set. Let
$\widetilde{p}_t^{C}$ and $\widetilde{q}_t^{I}$ denote the teacher and student distributions normalized over $\mathcal{K}_t$.
The on-policy distillation objective is defined as
\begin{equation}
\mathcal{L}_{\mathrm{OPD}}
=
\mathbb{E}_{\hat{\mathbf{s}}\sim
p_{\theta}(\cdot\mid H^{I})}
\left[
\frac{1}{|\hat{\mathbf{s}}|}
\sum_{t=1}^{|\hat{\mathbf{s}}|}
\operatorname{JSD}
\left(
\widetilde{p}_t^{C},
\widetilde{q}_t^{I}
\right)
\right],
\label{eq:item_opd}
\end{equation}
where $\operatorname{JSD}(\cdot,\cdot)$ denotes the Jensen--Shannon divergence.
In this way, the Crop Teacher provides search-target-centric supervision under the states that the student is likely to encounter during its own autoregressive generation.

\paragraph{Granularity-aware attention guidance.}
On-policy distillation does not explicitly regulate which visual regions the student relies on to make these predictions.
We therefore further guide the attention from SID tokens to visual tokens.

Let
$A_{t,v}^{(l,h)}$
denote the attention weight from the $t$-th SID token to visual token $v$ at layer $l$ and attention head $h$.
We first average the attention weights over all $N_h$ heads:
\begin{equation}
\bar{A}_{t,v}^{(l)}
=
\frac{1}{N_h}
\sum_{h=1}^{N_h}
A_{t,v}^{(l,h)}.
\label{eq:item_attention_head_average}
\end{equation}
We then retain the visual-token columns and normalize them to obtain a visual attention distribution:
\begin{equation}
\pi_t^{(l)}(v)
=
\frac{
\bar{A}_{t,v}^{(l)}
}{
\sum_{v'\in\mathcal{V}}
\bar{A}_{t,v'}^{(l)}
+\epsilon
},
\qquad
v\in\mathcal{V},
\label{eq:item_visual_attention}
\end{equation}
where $\mathcal{V}$ denotes the set of visual tokens.

The SID follows a coarse-to-fine semantic structure.
We exploit this property to guide both \emph{where} the model attends and \emph{how concentrated} its attention should be as SID generation proceeds.

% \paragraph{ROT attention guidance.}
% Let $M(v)\in\{0,1\}$ denote the visual-token-level ROT mask derived from the target-product bounding box.
% For the $t$-th SID token, we measure the amount of visual attention falling inside the target-product region as
% \begin{equation}
% m_t^{(l)}
% =
% \sum_{v\in\mathcal{V}}
% \pi_t^{(l)}(v)M(v).
% \label{eq:item_rot_mass}
% \end{equation}

\paragraph{ROT attention guidance.}
We define the bounding-box region of the search target as the \textbf{region of target (ROT)} and map it to the corresponding visual tokens to obtain a binary mask $M(v)\in\{0,1\}$.
For the $t$-th SID token, we measure the amount of visual attention falling within the ROT as
\begin{equation}
m_t^{(l)}
=
\sum_{v\in\mathcal{V}}
\pi_t^{(l)}(v)M(v).
\label{eq:item_rot_mass}
\end{equation}

We associate each SID token with its semantic granularity $g(t)$ and assign a granularity-dependent coefficient $\alpha_{g(t)}$.
In particular, increasingly fine-grained SID tokens receive progressively stronger ROT constraints:
\begin{equation}
\alpha_{\mathrm{coarse}}
<
\alpha_{\mathrm{medium}}
<
\alpha_{\mathrm{fine}},
\qquad
\alpha_{\mathrm{sep}}=0.
\label{eq:item_rot_weights}
\end{equation}
The ROT attention loss is defined as
\begin{equation}
\mathcal{L}_{\mathrm{rot}}
=
\operatorname{Mean}_{l,t}
\left[
-\alpha_{g(t)}
\log
\left(
m_t^{(l)}+\epsilon
\right)
\right].
\label{eq:item_rot_loss}
\end{equation}
This design imposes a relatively weak constraint when generating coarse SID tokens, while increasingly encouraging fine-grained predictions to rely on visual evidence inside the region of search target.

\paragraph{Coarse-to-fine entropy guidance.}
To regulate the concentration of visual attention at different SID granularities, we further introduce an entropy-based attention loss.
% Our design follows a coarse-to-fine principle, \textit{i.e.}, coarse SID tokens preserve a broader attention distribution to capture global visual characteristics, whereas medium- and fine-grained tokens should attend more selectively to discriminative visual details.
Our design follows a coarse-to-fine principle, with broader attention for coarse SID tokens and progressively concentrated attention for finer-grained tokens.
To this end, we define the attention entropy as
\begin{equation}
\mathcal{H}_t^{(l)}
=
-
\sum_{v\in\mathcal{V}}
\pi_t^{(l)}(v)
\log
\left(
\pi_t^{(l)}(v)+\epsilon
\right).
\label{eq:item_attention_entropy}
\end{equation}
We further introduce a granularity-dependent direction coefficient $\gamma_{g(t)}$ to control the desired entropy at each SID level as
\begin{equation}
\gamma_{\mathrm{coarse}}=+1,
\qquad
\gamma_{\mathrm{medium}}
=
\gamma_{\mathrm{fine}}
=-1,
\qquad
\gamma_{\mathrm{sep}}=0.
\label{eq:item_entropy_direction}
\end{equation}
The entropy objective is then
\begin{equation}
\mathcal{L}_{\mathrm{ent}}
=
\operatorname{Mean}_{l,t}
\left[
-\gamma_{g(t)}
\mathcal{H}_t^{(l)}
\right].
\label{eq:item_entropy_loss}
\end{equation}

% The ROT objective specifies where the visual evidence should come from, but does not control the spatial concentration of the attention distribution.
% We therefore further regularize its entropy:
% \begin{equation}
% \mathcal{H}_t^{(l)}
% =
% -
% \sum_{v\in\mathcal{V}}
% \pi_t^{(l)}(v)
% \log
% \left(
% \pi_t^{(l)}(v)+\epsilon
% \right).
% \label{eq:item_attention_entropy}
% \end{equation}

% We define a granularity-dependent direction coefficient $\gamma_{g(t)}$ such that

Minimizing Eq.~\eqref{eq:item_entropy_loss} encourages coarse SID tokens to maintain a relatively high-entropy visual attention distribution and capture broader visual evidence, whereas medium- and fine-grained SID tokens are encouraged to progressively concentrate their attention on discriminative product details.
Therefore, the ROT and entropy objectives are complementary: the former determines \emph{where} the model should gather visual evidence, while the latter regulates \emph{how broadly or selectively} the evidence should be collected at different semantic granularities.

\paragraph{Training objective.}
Overall, the training objective is
\begin{equation}
\mathcal{L}_{\mathrm{item}}
=
\mathcal{L}_{\mathrm{CE}}
+
\lambda_{\mathrm{opd}}
\mathcal{L}_{\mathrm{OPD}}
+
\lambda_{\mathrm{rot}}
\mathcal{L}_{\mathrm{rot}}
+
\lambda_{\mathrm{ent}}
\mathcal{L}_{\mathrm{ent}}.
\label{eq:item_total_loss}
\end{equation}
% where $\mathcal{L}_{\mathrm{CE}}$ ensures correct SID generation,
% $\mathcal{L}_{\mathrm{OPD}}$ transfers product-centered predictive behavior from the Crop Teacher,
% and $\mathcal{L}_{\mathrm{rot}}$ and $\mathcal{L}_{\mathrm{ent}}$ jointly guide the visual evidence used during coarse-to-fine SID decoding.
Minimizing Eq.~\eqref{eq:item_entropy_loss} enforces the desired coarse-to-fine attention concentration across SID granularities.
The objective enable the student to internalize target-focused perception while operating solely on the original query image.

% \vspace{-1em}
\subsection{Selective Auxiliary-Evidence Utilization Mechanism}
\label{sec:item_auxiliary}

After establishing an target-centric representation, the model should further exploit auxiliary evidence, while avoiding irrelevant or misleading information.
To this end, we introduce \textbf{Selective Auxiliary-Evidence Utilization Mechanism}, which consists of an auxiliary enhancer  for identifying and enhancing useful auxiliary information, and a selective distillation objective for determining which auxiliary capability should be internalized.

\subsubsection{Auxiliary Enhancer }
\label{sec:auxiliary_router}

Given the crop tokens
$H^{I}=\{h_i^{I}\}_{i=1}^{N}$,
we summarize the tokens into an \textbf{Target Anchor},
\begin{equation}
c^{I}
=
\frac{
\sum_{i=1}^{N} a_i^{I} h_i^{I}
}{
\sum_{i=1}^{N} a_i^{I}+\epsilon
}.
\label{eq:item_anchor}
\end{equation}
% where $a_i^{I}$ is the item relevance score obtained from the Target Enhancer.

The auxiliary enhancer  evaluates whether a visual token provides complementary information for the current target.
Specifically, we compute the auxiliary relevance score by jointly considering the local visual token and the Target Anchor:
\begin{equation}
a_i^{A}
=
\sigma
\left(
g_A(h_i^{I},c^{I})
\right),
\label{eq:auxiliary_score}
\end{equation}
where
\begin{equation}
g_A(h_i^{I},c^{I})
=
w_A^{\top}
\operatorname{GELU}
\left(
W_A
\left[
\operatorname{LN}(h_i^{I});
\operatorname{LN}(c^{I});
\operatorname{LN}(h_i^{I})
\odot
\operatorname{LN}(c^{I})
\right]
\right).
\label{eq:auxiliary_router}
\end{equation}

The selected auxiliary visual tokens are further enhanced through \textbf{Token-level Residual Modulation}, \textit{i.e.},
\begin{equation}
h_i^{I+A}
=
h_i^{I}
+
\beta
\lambda_A
a_i^{A}
F_A(h_i^{I}),
\label{eq:auxiliary_residual}
\end{equation}
where $F_A(\cdot)$ denotes a learnable residual transformation, $\lambda_A$ controls the residual magnitude, and $\beta$ controls the overall contribution of auxiliary information.
The crop+OCR tokens are computed as
\begin{equation}
H^{I+A}
=
\{h_i^{I+A}\}_{i=1}^{N}.
\label{eq:item_auxiliary_representation}
\end{equation}

\subsubsection{Selective Auxiliary Distillation Objective}
\label{sec:selective_auxiliary_distillation}

OCR cues can be useful, irrelevant, or even misleading for retrieval.
Moreover, even useful auxiliary information may not always be reliably captured by the student from the original query image.
We therefore determine auxiliary supervision from two perspectives: whether the auxiliary information is \emph{useful} according to the teacher, and whether it is \emph{accessible} to the student.

We employ two teacher models.
The \textbf{Crop Teacher} takes only the cropped image as input, while the \textbf{OCR Teacher} additionally receives the pre-extracted OCR information.
Let $p_t^{C}$ and $p_t^{O}$ denote their SID distributions at decoding step $t$.
The utility of auxiliary information is measured by the improvement of the OCR Teacher over the Crop Teacher on the ground-truth SID token:
\begin{equation}
u_t^{T}
=
\log p_t^{O}(s_t^{*})
-
\log p_t^{C}(s_t^{*}).
\label{eq:teacher_utility}
\end{equation}
A larger $u_t^{T}$ indicates that the auxiliary information provides additional evidence for predicting the target SID.

We further evaluate whether such auxiliary information can be exploited by the student.
Let
\begin{equation}
q_t^{I}(\cdot)
=
p_{\theta}
\left(
\cdot
\mid
H^{I},
\mathbf{s}_{<t}
\right),
\end{equation}
and
\begin{equation}
q_t^{I+A}(\cdot)
=
p_{\theta}
\left(
\cdot
\mid
H^{I+A},
\mathbf{s}_{<t}
\right)
\end{equation}
denote the student predictions using crop tokens $H^I$ and crop+OCR tokens $H^{I+A}$, respectively.
We define the accessibility as
\begin{equation}
u_t^{S}
=
\log q_t^{I+A}(s_t^{*})
-
\log q_t^{I}(s_t^{*}).
\label{eq:student_accessibility}
\end{equation}

Based on these two signals, we construct soft selection weights:
\begin{equation}
w_t^{\mathrm{help}}
=
\sigma
\left(
\tau_T u_t^{T}-\rho_T
\right),
\qquad
w_t^{\mathrm{cap}}
=
\sigma
\left(
\tau_S u_t^{S}-\rho_S
\right),
\label{eq:selective_weights}
\end{equation}
where $\rho_T$ and $\rho_S$ are the selection thresholds for auxiliary utility and student accessibility, respectively, while $\tau_T$ and $\tau_S$ control the sharpness of the corresponding soft gates.
Here, $w_t^{\mathrm{help}}$ measures whether the auxiliary information is beneficial, whereas $w_t^{\mathrm{cap}}$ measures whether it is accessible to the student.
% where $w_t^{\mathrm{help}}$ measures whether the auxiliary information is beneficial, while $w_t^{\mathrm{cap}}$ measures whether such information can be exploited by the student.

When the auxiliary evidence is both useful and accessible, we encourage the auxiliary-enhanced student to approach the OCR Teacher and explicitly realize a prediction gain over the target-only representation:
\begin{equation}
\mathcal{L}_{\mathrm{open}}
=
\mathcal{L}_{\mathrm{contrast}}
+
\lambda_{\mathrm{gain}}
\mathcal{L}_{\mathrm{gain}},
\label{eq:open_loss}
\end{equation}
where
\begin{equation}
\mathcal{L}_{\mathrm{contrast}}
=
-
\log
\frac{
\exp
\left(
s(q_t^{I+A},p_t^{O})/\tau
\right)
}{
\exp
\left(
s(q_t^{I+A},p_t^{O})/\tau
\right)
+
\exp
\left(
s(q_t^{I+A},p_t^{C})/\tau
\right)
},
\label{eq:contrastive_distillation}
\end{equation}
and
\begin{equation}
\mathcal{L}_{\mathrm{gain}}
=
\left[
m
-
\left(
\log q_t^{I+A}(s_t^{*})
-
\log q_t^{I}(s_t^{*})
\right)
\right]_{+}.
\label{eq:gain_loss}
\end{equation}

In contrast, when the auxiliary information is not useful, we suppress its influence by encouraging the auxiliary-enhanced prediction to remain close to the target-only prediction and the Crop Teacher:
\begin{equation}
\mathcal{L}_{\mathrm{close}}
=
\operatorname{JSD}
\left(
q_t^{I+A},
q_t^{I}
\right)
+
\lambda_C
\operatorname{JSD}
\left(
q_t^{I+A},
p_t^{C}
\right).
\label{eq:close_loss}
\end{equation}

When the auxiliary evidence is useful but not yet accessible, we apply neither transfer nor suppression, avoiding forced imitation of inaccessible capability while allowing the student to improve during training.

Regardless of auxiliary-evidence utility and accessibility, we apply ground-truth SID supervision and align the target-only prediction with the Crop Teacher to keep SID generation centered on the search target.
We define a shared base objective as
\begin{equation}
\mathcal{L}_{\mathrm{base}}
=
\operatorname{CE}
\left(
q_t^{I},
s_t^{*}
\right)
+
\operatorname{CE}
\left(
q_t^{I+A},
s_t^{*}
\right)
+
\lambda_I
\operatorname{JSD}
\left(
q_t^{I},
p_t^{C}
\right),
\label{eq:auxiliary_base_loss}
\end{equation}
where $\operatorname{CE}(\cdot,\cdot)$ denotes the cross-entropy loss.
The overall objective is computed as 
\begin{equation}
\mathcal{L}_{t}
=
\mathcal{L}_{\mathrm{base}}
+
\lambda_{\mathrm{gate}}
\left(w_t^{\mathrm{help}}
w_t^{\mathrm{cap}}
\mathcal{L}_{\mathrm{open}}
+
\left(
1-w_t^{\mathrm{help}}
\right)
\mathcal{L}_{\mathrm{close}}\right),
\label{eq:selective_auxiliary_objective}
\end{equation}
where $\lambda_{\mathrm{gate}}$ controls the overall strength of the selectively gated auxiliary-evidence objectives.
The objective transfers auxiliary capability only when the additional evidence is both beneficial and accessible to the student, while suppressing irrelevant or misleading auxiliary information.

% The overall objective is computed as 
% \begin{equation}
% \mathcal{L}_{t}
% =
% \mathcal{L}_{\mathrm{base}}
% +
% w_t^{\mathrm{help}}
% w_t^{\mathrm{cap}}
% \mathcal{L}_{\mathrm{open}}
% +
% \left(
% 1-w_t^{\mathrm{help}}
% \right)
% \mathcal{L}_{\mathrm{close}},
% \label{eq:selective_auxiliary_objective}
% \end{equation}
% where  $\mathcal{L}_{\mathrm{base}}$ is computed as
% \begin{equation}
% \mathcal{L}_{\mathrm{base}}
% =
% \operatorname{CE}
% \left(
% q_t^{I},
% s_t^{*}
% \right)
% +
% \operatorname{CE}
% \left(
% q_t^{I+A},
% s_t^{*}
% \right)
% +
% \lambda_I
% \operatorname{JSD}
% \left(
% q_t^{I},
% p_t^{C}
% \right),
% \label{eq:auxiliary_base_loss}
% \end{equation}

% \vspace{-1em}
\subsection{Generative Inference}

\label{sec:inference}

After training, both target focusing and auxiliary-evidence utilization are internalized into the generative retriever.
Given only the original query image, the target enhancer identifies and enhances visual tokens of search target, enabling \textit{Zooming without Cropping}.
Conditioned on the resulting target information, the auxiliary enhancer  further identifies and enhances useful auxiliary cues directly from the visual tokens, enabling \textit{Reading without OCR}.
Meanwhile, the learned target-focused perception encourages SID decoding to remain centered on visual tokens of search target, with auxiliary cues serving only as complementary evidence for fine-grained identification.
The resulting enhanced visual tokens are directly used for SID generation, while the crop and OCR teachers are completely removed during inference.

\section{Data Construction}
We construct our dataset from online image-search logs collected from a large-scale e-commerce platform: Pailitao.
For training, we retain query images associated with behavioral SIDs from clicks, purchases, add-to-cart, and favorite actions, and require each query to have more than three behavioral SIDs.
We then select the seven most frequent subcategories: women's clothing, children's clothing, men's clothing, women's shoes, men's shoes, digital products, and furniture.
The test set is collected from a temporally disjoint period to avoid query-image overlap with training, and further requires more than three behavioral SIDs and at least one purchase behavior.
All test samples are further verified by both human annotators and matching models to ensure that each query image and its associated item depict the same product.
We evaluate retrieval performance based on click and purchase behaviors.
We use the same seven subcategories as in the training set for evaluation.

The training and test sets contain 1,159,746 and 8,647 query images, respectively, covering seven product categories.
The training set contains 275,161 women's shoes, 232,529 furniture, 200,000 women's clothing, 163,526 men's clothing, 144,629 children's clothing, 100,628 3C products, and 43,273 men's shoes queries, while the corresponding numbers in the test set are 1,095, 483, 4,258, 624, 1,420, 576, and 191.
Each query is associated with multiple positive items: the average numbers are 5.87 and 5.61 for training and test, respectively, with a median of 5 for both sets.
Most queries contain 4--6 positive items, accounting for approximately 57\% of the data.
% Our training and test sets contain XXX and XXX query images, respectively.
% The category distributions of the two sets are shown in Fig.~\ref{fig:data_statistics}, covering seven major categories.
% We further report the distribution of the number of positive samples associated with each query in Fig.~\ref{fig:data_statistics}.
The training and test sets will be publicly released upon publication of the paper.

% We further report the statistics of the constructed training and test sets in Fig.~\ref{fig:data_statistics}.
% The figure presents the query distribution across the seven selected categories, the distribution of behavioral SIDs associated with each query, and the distributions of click and purchase SIDs in the test set.
% The training and test sets contain XXX and XXX queries, covering XXX and XXX unique SIDs, respectively.

\section{Experiments}

\subsection{Settings}
% \textbf{Data Construction.}

% For the training set, we retain query images associated with behavioral SIDs, where user behaviors include clicks, purchases, adding to cart, and adding to favorites.
% To ensure reliable behavioral supervision, we further require each query to be associated with more than three behavioral SIDs.
% From the resulting data, we select the seven most frequent subcategories, including women's clothing, children's clothing, men's clothing, women's shoes, men's shoes, digital products, and furniture.
% We construct the test set from online logs collected during a time period completely disjoint from that of the training set, ensuring that the query images in the two sets do not overlap.
% We apply a stricter filtering criterion for evaluation: each query should be associated with more than three behavioral SIDs and must contain at least one purchase behavior.
% For evaluation, we consider two types of user behaviors in the test set: clicks and purchases.

% \small

\begin{table*}[t]
\centering
\caption{Comparison of retrieval performance (\%) on the test set.}
% \vspace{-1em}
\label{tab:main_results}
\resizebox{\textwidth}{!}{
\begin{tabular}{llcccccccccc}
\toprule
\multirow{2}{*}{Method}
& \multirow{2}{*}{Input}
& \multicolumn{5}{c}{CLICK}
& \multicolumn{5}{c}{PURCHASE} \\
\cmidrule(lr){3-7}
\cmidrule(lr){8-12}
& & H@1 & H@5 & H@10 & R@5 & R@10
& H@1 & H@5 & H@10 & R@5 & R@10 \\
\midrule

\multicolumn{12}{l}{\textbf{Traditional Retrieval}} \\
DINOv3 & Crop 
& 36.38 & 58.61 & 64.36 & 32.60 & 39.87
& 14.64 & 36.98 & 44.35 & 35.84 & 43.27 \\
CLIP & Crop
& 30.57 & 53.90  & 61.69 & 28.83 & 36.73
& 12.24 & 32.57 & 41.11 & 31.54 & 40.05 \\
\midrule

\multicolumn{12}{l}{\textbf{Existing Generative Retrieval}} \\
IRGen & Crop
& 29.72 & 55.02 & 60.06 & 28.87 & 36.03
& 11.54 & 32.81 & 39.85 & 32.00 & 39.14 \\
GENIUS & Crop
& 24.62 & 42.12 & 45.19 & 21.99 & 26.93
& 9.37 & 25.03 & 29.72 & 24.37 & 29.28 \\
\midrule

\multicolumn{12}{l}{\textbf{Teacher Models}} \\
Crop Teacher & Crop
& 37.48 & 68.16 & 74.25 & 39.02 & 50.41
& 14.70 & 44.28 & 54.47 & 43.41 & 54.02 \\
OCR Teacher & Crop + OCR
& 39.74 & 70.77 & 77.02 & 41.22 & 52.88
& 16.00 & 46.47 & 57.06 & 45.54 & 56.48 \\
\midrule

\multicolumn{12}{l}{\textbf{Full-Image Models}} \\
SFT Baseline & Full Image
& 29.25 & 56.13 & 63.24 & 30.71 & 40.74
& 11.55 & 34.69 & 44.57 & 34.06 & 44.05 \\
\textbf{Ours} & Full Image
& \textbf{41.88} & \textbf{73.33} & \textbf{79.19} & \textbf{43.46} & \textbf{55.27}
& \textbf{17.20} & \textbf{49.39} & \textbf{60.12} & \textbf{48.46} & \textbf{59.48} \\
\bottomrule
% \multicolumn{12}{l}{\textbf{Student Models}} \\
% SFT Baseline & Full Image
% & 29.25 & 56.13 & 63.24 & 30.71 & 40.74
% & 11.55 & 34.69 & 44.57 & 34.06 & 44.05 \\
% \textbf{Ours} & Full Image
% & \textbf{41.88} & \textbf{73.33} & \textbf{79.19} & \textbf{43.46} & \textbf{55.27}
% & \textbf{17.20} & \textbf{49.39} & \textbf{60.12} & \textbf{48.46} & \textbf{59.48} \\
\end{tabular}
}
\end{table*}

\paragraph{Evaluation Metrics.}
We adopt two metrics, R@K and H@K, to evaluate retrieval performance.
R@K measures the proportion of behavior-associated items that are successfully retrieved among the top-$K$ results, while H@K measures whether at least one behavior-associated SID is retrieved.
Formally, given the ground-truth behavior-associated SID set $\mathcal{S}^{\mathrm{beh}}$ and the top-$K$ retrieved SID set $\hat{\mathcal{S}}_K$,
\begin{equation}
\mathrm{R@K}
=
\frac{
\left|
\hat{\mathcal{S}}_K
\cap
\mathcal{S}^{\mathrm{beh}}
\right|
}{
\min
\left(
K,
\left|\mathcal{S}^{\mathrm{beh}}\right|
\right)
},
\end{equation}
and
\begin{equation}
\mathrm{H@K}
=
\mathbb{I}
\left(
\left|
\hat{\mathcal{S}}_K
\cap
\mathcal{S}^{\mathrm{beh}}
\right|
>0
\right),
\end{equation}
where $\mathbb{I}(\cdot)$ denotes the indicator function.
For both metrics, we separately compute the results based on click and purchase behaviors.

\paragraph{Comparison Methods.}
We compare our method with two representative generative image retrieval methods, IRGen~\cite{zhang2024irgen} and GENIUS~\cite{kim2025genius}, and follow the model configurations used in their original papers.
Specifically, GENIUS adopts CLIP ViT-L/14 as the visual encoder and a 6-layer T5 decoder, while IRGen employs CLIP ViT-B/16 as the visual encoder and a 24-layer Transformer decoder.
% Both methods are initialized from scratch and trained on our constructed training set.
Both methods are initialized from corresponding pretrained models and trained on our constructed training set.
We additionally compare with two conventional similarity-based retrieval methods, DINOv3~\cite{simeoni2025dinov3} and CLIP~\cite{radford2021learning}, both using ViT-B/16.
Their released pretrained weights are used for initialization, followed by fine-tuning on our training set.
All trained model checkpoints will be publicly released upon acceptance of the paper.

% Detailed architectures and implementation settings are provided in the supplementary material.

\paragraph{Implementation Details.}
We adopt Qwen3.5-0.8b~\cite{yang2025qwen3} as our base model.
The SID consists of three codebook levels, each with a codebook size of 8,192.
For the teacher models, the \textbf{Crop Teacher} takes the cropped search target as input, while the \textbf{OCR Teacher} takes both the cropped target image and the pre-extracted OCR as input.
Both teachers are trained with SID supervision.
Our student model takes only the original query image as input.
% More details, such as  hyperparameters, are presented in supplementary material.
We train the model with a batch size of 1,024 and an initial learning rate of $1\times10^{-4}$, which is decayed using a cosine schedule.

We set $\lambda_{\mathrm{opd}}=2.0$, $\lambda_{\mathrm{roi}}=0.1$, and $\lambda_{\mathrm{ent}}=0.02$.
For on-policy distillation, we compute the forward KL divergence over the teacher's top-100 predictions.
The weights for ROT attention guidance are set to $\alpha=(0.1, 0.3, 0.6, 0)$ for coarse, medium, fine, and separator SID tokens, respectively.
We set
$\tau_T=\tau_S=1.0$,
$\rho_T=\rho_S=0.0$,
$\lambda_{\mathrm{gain}}=0.5$,
$m=0.5$,
$\lambda_C=0.5$,
and $\lambda_{\mathrm{gate}}=0.02$.
% Here, $\lambda_{\mathrm{gate}}$ controls the overall strength of the selectively gated auxiliary-evidence objectives.
% Additional implementation details and hyperparameter settings are provided in the supplementary material.

% \vspace{-1em}
\subsection{Main Results}
\label{sec:main_results}

% We first compare our model with two existing generative image retrieval methods, IRGen and GENIUS.
% Table~\ref{tab:main_results} reports the main results.
% Under the same crop-based setting, our model consistently outperforms IRGen and GENIUS, while IRGen outperforms GENIUS.
% We attribute this trend to decoder capacity, with decoder size following Qwen3.5-0.8B > IRGen > GENIUS, enabling stronger SID modeling and better retrieval performance.
% We attribute this trend to decoder capacity, as the larger decoder of Qwen3.5-0.8B better supports SID modeling and generation than those of IRGen and GENIUS.

\begin{table*}[t]
\centering
\caption{
Retrieval performance (\%) under different search-target ROT area ratios.
$\Delta$ denotes the absolute improvement of our method over the SFT Baseline.
}
% \vspace{-1em}
\label{tab:rot_analysis}
\setlength{\tabcolsep}{3.8pt}
\resizebox{\textwidth}{!}{
\begin{tabular}{llcccccccccc}
\toprule
\multirow{2}{*}{ROT Ratio}
& \multirow{2}{*}{Method}
& \multicolumn{5}{c}{CLICK}
& \multicolumn{5}{c}{PURCHASE} \\
\cmidrule(lr){3-7}
\cmidrule(lr){8-12}
& & H@1 & H@5 & H@10 & R@5 & R@10
& H@1 & H@5 & H@10 & R@5 & R@10 \\
\midrule

\multirow{3}{*}{0--5\%}
& SFT Baseline
& 18.01 & 36.53 & 42.42 & 19.12 & 26.44
& 6.23 & 20.71 & 29.12 & 20.51 & 28.85 \\
& Ours
& 36.53 & 65.49 & 72.05 & 37.82 & 49.92
& 15.32 & 42.59 & 54.38 & 41.83 & 53.85 \\
& $\Delta$
& \textbf{+18.52} & \textbf{+28.96} & \textbf{+29.63}
& \textbf{+18.71} & \textbf{+23.49}
& \textbf{+9.09} & \textbf{+21.89} & \textbf{+25.25}
& \textbf{+21.31} & \textbf{+25.00} \\
\midrule

\multirow{3}{*}{5--10\%}
& SFT Baseline
& 26.02 & 50.52 & 57.25 & 26.86 & 35.88
& 9.85 & 29.06 & 38.11 & 28.71 & 37.81 \\
& Ours
& 42.11 & 75.18 & 81.18 & 45.20 & 57.25
& 17.29 & 50.68 & 62.05 & 49.66 & 61.30 \\
& $\Delta$
& \textbf{+16.09} & \textbf{+24.66} & \textbf{+23.94}
& \textbf{+18.34} & \textbf{+21.36}
& \textbf{+7.45} & \textbf{+21.62} & \textbf{+23.94}
& \textbf{+20.95} & \textbf{+23.49} \\
\midrule

\multirow{3}{*}{10--20\%}
& SFT Baseline
& 29.59 & 57.88 & 64.93 & 32.04 & 42.36
& 11.51 & 37.04 & 47.30 & 36.40 & 47.08 \\
& Ours
& 43.82 & 77.20 & 82.46 & 46.98 & 58.88
& 18.43 & 54.35 & 65.02 & 53.60 & 64.28 \\
& $\Delta$
& \textbf{+14.23} & \textbf{+19.32} & \textbf{+17.54}
& \textbf{+14.94} & \textbf{+16.52}
& \textbf{+6.92} & \textbf{+17.31} & \textbf{+17.72}
& \textbf{+17.20} & \textbf{+17.20} \\
\midrule

\multirow{3}{*}{20--40\%}
& SFT Baseline
& 32.76 & 61.20 & 68.62 & 33.99 & 44.70
& 13.39 & 38.80 & 49.12 & 38.15 & 48.49 \\
& Ours
& 43.84 & 75.18 & 80.88 & 44.44 & 56.64
& 18.09 & 50.95 & 61.75 & 49.82 & 61.11 \\
& $\Delta$
& \textbf{+11.08} & \textbf{+13.98} & \textbf{+12.25}
& \textbf{+10.45} & \textbf{+11.94}
& \textbf{+4.69} & \textbf{+12.15} & \textbf{+12.63}
& \textbf{+11.67} & \textbf{+12.62} \\
\midrule

\multirow{3}{*}{$\geq$40\%}
& SFT Baseline
& 28.93 & 56.06 & 63.39 & 30.12 & 40.19
& 11.52 & 33.55 & 43.28 & 32.78 & 42.53 \\
& Ours
& 37.39 & 66.75 & 73.65 & 37.92 & 48.76
& 14.59 & 41.60 & 51.86 & 40.82 & 51.43 \\
& $\Delta$
& \textbf{+8.46} & \textbf{+10.68} & \textbf{+10.26}
& \textbf{+7.80} & \textbf{+8.58}
& \textbf{+3.06} & \textbf{+8.04} & \textbf{+8.58}
& \textbf{+8.04} & \textbf{+8.89} \\
\bottomrule
\end{tabular}
}
\end{table*}

Table~\ref{tab:main_results} reports the main results.
Our method outperforms all existing methods, achieving an average improvement of 12.16\% over the second-best method across the ten metrics, demonstrating its superior retrieval performance.
Among generative retrieval methods under the same crop-based setting, we observe that our model consistently outperforms IRGen and GENIUS, while IRGen further outperforms GENIUS.
We attribute this trend to decoder capacity, with decoder size following Qwen3.5-0.8B $>$ IRGen $>$ GENIUS, leading to stronger SID modeling and better retrieval performance.

More importantly, our model outperforms the directly SFT-trained baseline by an average of 13.8\% across the ten metrics, demonstrating the effectiveness of the proposed capability internalization.
It also surpasses the Crop Teacher and OCR Teacher by 4.76\% and 2.76\% across the ten metrics, respectively, although they use Crop and Crop+OCR inputs during both training and inference.
These results demonstrate that our method effectively internalizes target-focusing and auxiliary-evidence utilization capabilities, achieving \textit{Zooming without Cropping} and \textit{Reading without OCR} using only the original query image.
Moreover, outperforming the OCR Teacher suggests that selective internalization enables the model to exploit beneficial auxiliary cues while avoiding indiscriminate reliance on them.

\subsection{Target-Focusing Capability Analysis}
\label{sec:item_focusing_analysis}

To verify whether our method effectively internalizes the target-focusing capability, we analyze model performance under different sizes of search target.
% Specifically, we define the search-target bounding box as the region of target (ROT) and group test queries by the number of visual tokens covered by the ROT.
Specifically, we use the bounding box of the search target in each query image as its region of target (ROT), and divide the test queries into different groups according to the number of visual tokens covered by the ROT. 
We then compare our method with the directly SFT baseline across these groups.
The results are shown in Table~\ref{tab:rot_analysis}.
Our method consistently improves performance across different ROT groups, with more pronounced gains when the search target occupies a smaller region of the query image.
These results demonstrate that our method can focus on the search target in complex scenes, validating its \textit{Zooming without Cropping} capability.

\begin{table*}[t]
\centering
\caption{
Retrieval performance (\%) on queries with and without OCR information.
$\Delta_{\mathrm{SFT}}$ and $\Delta_{\mathrm{Crop}}$ denote the absolute improvements of our method over the SFT Baseline and Crop Teacher, respectively.
}
% \vspace{-1em}
\label{tab:ocr_analysis}
\setlength{\tabcolsep}{3.8pt}
\resizebox{\textwidth}{!}{
\begin{tabular}{llcccccccccc}
\toprule
\multirow{2}{*}{OCR}
& \multirow{2}{*}{Method}
& \multicolumn{5}{c}{CLICK}
& \multicolumn{5}{c}{PURCHASE} \\
\cmidrule(lr){3-7}
\cmidrule(lr){8-12}
& & H@1 & H@5 & H@10 & R@5 & R@10
& H@1 & H@5 & H@10 & R@5 & R@10 \\
\midrule

\multirow{5}{*}{With OCR}
& SFT Baseline
& 24.50 & 49.84 & 57.40 & 25.69 & 34.87
& 9.45 & 28.76 & 38.04 & 27.87 & 37.20 \\
& Crop Teacher
& 32.95 & 63.66 & 70.07 & 34.88 & 45.61
& 12.53 & 39.96 & 49.76 & 38.92 & 48.93 \\
& Ours
& 37.95 & 70.84 & 77.22 & 40.56 & 52.19
& 15.37 & 46.54 & 57.71 & 45.46 & 56.73 \\
& $\Delta_{\mathrm{SFT}}$
& \textbf{+13.44} & \textbf{+21.00} & \textbf{+19.82}
& \textbf{+14.87} & \textbf{+17.32}
& \textbf{+5.92} & \textbf{+17.78} & \textbf{+19.68}
& \textbf{+17.59} & \textbf{+19.53} \\
& $\Delta_{\mathrm{Crop}}$
& \textbf{+5.00} & \textbf{+7.18} & \textbf{+7.15}
& \textbf{+5.67} & \textbf{+6.58}
& \textbf{+2.84} & \textbf{+6.58} & \textbf{+7.96}
& \textbf{+6.54} & \textbf{+7.80} \\
\midrule

\multirow{5}{*}{Without OCR}
& SFT Baseline
& 32.38 & 60.34 & 67.13 & 34.08 & 44.66
& 12.95 & 38.68 & 48.95 & 38.21 & 48.65 \\
& Crop Teacher
& 40.53 & 71.20 & 77.06 & 41.81 & 53.65
& 16.16 & 47.19 & 57.65 & 46.47 & 57.48 \\
& Ours
& 44.46 & 75.15 & 80.76 & 45.50 & 57.47
& 18.43 & 51.36 & 61.92 & 50.53 & 61.51 \\
& $\Delta_{\mathrm{SFT}}$
& \textbf{+12.08} & \textbf{+14.81} & \textbf{+13.63}
& \textbf{+11.42} & \textbf{+12.81}
& \textbf{+5.48} & \textbf{+12.68} & \textbf{+12.97}
& \textbf{+12.31} & \textbf{+12.86} \\
& $\Delta_{\mathrm{Crop}}$
& \textbf{+3.93} & \textbf{+3.95} & \textbf{+3.70}
& \textbf{+3.70} & \textbf{+3.82}
& \textbf{+2.26} & \textbf{+4.16} & \textbf{+4.28}
& \textbf{+4.06} & \textbf{+4.04} \\
\bottomrule
\end{tabular}
}
\end{table*}

\subsection{OCR Reading Capability Analysis}
\label{sec:ocr_analysis}

To verify whether our method can effectively internalize the capability of utilizing textual cues, we divide the test queries into two groups according to whether OCR information is present in the query image.
% To evaluate whether our method effectively internalizes textual-cue utilization, we divide the test queries into two groups based on the presence of OCR information in the query image.
We then compare the performance of our method with the SFT baseline and the Crop Teacher.
The results are shown in Table~\ref{tab:ocr_analysis}.
Our method achieves more evident improvements on queries containing OCR, demonstrating its ability to exploit useful textual cues directly from the original image without explicit OCR input.
The performance on queries without OCR remains competitive, indicating that the model does not indiscriminately rely on textual cues.
These results validate both the \textit{Reading without OCR} capability and the effectiveness of selective auxiliary-evidence utilization.

% \vspace{-1em}
\subsection{Visualization}
\label{sec:attention_visualization}

\begin{figure}[t]
    \centering
        \includegraphics[width=\linewidth]{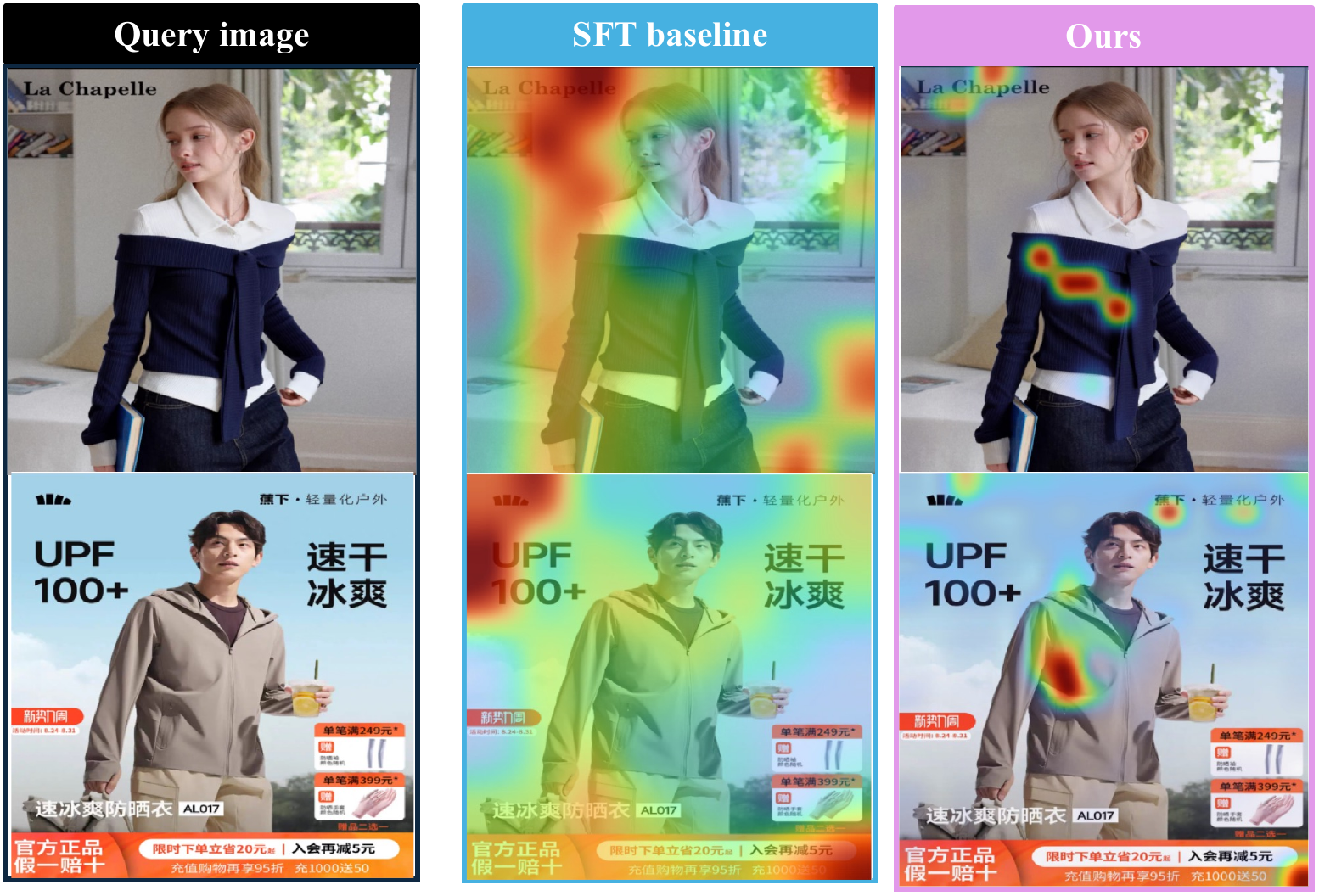}
    \caption{Visualization of answer-to-visual attention.
From left to right: query image, attention map of the SFT Baseline, and attention map of our model.}
    \label{fig:attention_visualization}
\end{figure}

To further analyze whether our model effectively focuses on the target during SID generation, we visualize the attention from SID tokens to visual tokens.
Following the same attention computation used in our training objective, we average the attention across heads and Transformer layers, aggregate the three SID levels, and project the resulting attention distribution back to the original image.
As shown in Fig.~\ref{fig:attention_visualization}, the SFT baseline exhibits relatively dispersed attention and is easily affected by background regions, whereas our model consistently concentrates more attention on the search-target region.
Moreover, we observe that our model can also attend to auxiliary textual cues in the query image without explicit OCR input.
These results demonstrate that our method effectively guides SID decoding to focus on search targets and useful auxiliary evidence, validating the effectiveness of target-focused perception mechanism and selective auxiliary-evidence utilization mechanism   in the proposed PailitaoGR method.
% These  results demonstrate that our method effectively guides SID decoding toward the core regions of search target, validating the proposed target-focusing mechanism.

% On queries with small ROTs, the attention mass inside the ROT increases from 8.2\% for the SFT baseline to 55.4\% for our model.

% \vspace{-1em}
\subsection{Ablation Studies}

% \paragraph{Effect of Core Components.}

\paragraph{Effect of Core Components.}
We progressively introduce target-focused perception mechanism and selective auxiliary-evidence utilization mechanism into the SFT Baseline.
As shown in Table~\ref{tab:component_ablation}, both components consistently improve retrieval performance, validating the effectiveness of search-target focusing and auxiliary-evidence utilization, respectively.

% \paragraph{Ablation of target-focused perception mechanism.}
% We further ablate the training objectives for target-focused perception mechanism, including on-policy distillation, ROT attention loss, and entropy regularization.
% Removing any of these objectives degrades performance, demonstrating their complementary roles in learning search-target-centered and coarse-to-fine visual attention.

\begin{table}[t]
\centering
\caption{Ablation study of the core components in PailitaoGR.}
\label{tab:component_ablation}
% \vspace{-1em}
\resizebox{\linewidth}{!}{
\begin{tabular}{lcc|cc|cc}
\toprule
\multirow{2}{*}{Method}
& \multicolumn{2}{c|}{Components}
& \multicolumn{2}{c|}{CLICK}
& \multicolumn{2}{c}{PURCHASE} \\
& Target & Auxiliary
& H@10 & R@10
& H@10 & R@10 \\
\midrule
SFT Baseline
&  & 
& 63.24 & 40.74
& 44.57 & 44.05 \\

+ Focusing Internalization
& \checkmark &
& 76.53 & 52.93
& 57.33 & 56.75 \\

PailitaoGR
& \checkmark & \checkmark
& \textbf{79.19} & \textbf{55.27}
& \textbf{60.12} & \textbf{59.48} \\
\bottomrule
\end{tabular}}
\end{table}

\paragraph{Ablation of target-focused perception mechanism.}
We conduct the ablation study for target-focused perception mechanism.
We ablate the three training objectives, including on-policy distillation, ROT- and entropy-based loss, as shown in Table~\ref{tab:item_ablation}.
The full model consistently achieves the best performance, confirming that the three objectives provide complementary supervision for target-focused perception.
Among them, OPD contributes the largest gains, highlighting the importance of transferring target-focused prediction behavior from the Crop Teacher.
ROT attention guidance also yields substantial gains, showing that explicit search-target region supervision complements the implicit target-focused behavior learned through OPD. Entropy regularization further improves performance by encouraging coarse-to-fine attention concentration.
% ROT attention guidance further provides strong spatial supervision, while entropy regularization brings additional improvements by encouraging coarse-to-fine attention concentration.
% Removing any objective degrades retrieval performance, demonstrating their complementary contributions to search-target focusing and coarse-to-fine attention learning.

\begin{table}
\centering
\caption{Ablation of target-focused perception mechanism.}
\label{tab:item_ablation}
\resizebox{\linewidth}{!}{
\begin{tabular}{lccc|cc|cc}
\toprule
\multirow{2}{*}{Method}
& \multicolumn{3}{c|}{Objective}
& \multicolumn{2}{c|}{CLICK}
& \multicolumn{2}{c}{PURCHASE} \\
& OPD & ROT & Entropy
& H@10 & R@10
& H@10 & R@10 \\
\midrule
w/o OPD
&  & \checkmark & \checkmark
& 73.05 & 49.40
& 53.94  & 53.25 \\

w/o ROT
& \checkmark &  & \checkmark
& 73.47 & 50.22
& 54.91 & 54.27 \\

w/o Entropy
& \checkmark & \checkmark &
& 75.64 & 52.07
&  56.72  & 55.98 \\

Full
& \checkmark & \checkmark & \checkmark
& \textbf{76.51} & \textbf{52.93}
& \textbf{57.52} & \textbf{56.86} \\
\bottomrule
\end{tabular}}
\end{table}

\paragraph{Ablation of selective auxiliary-evidence utilization mechanism.}
We further conduct the auxiliary-evidence ablation.
Starting from indiscriminate auxiliary transfer, we progressively introduce the utility and accessibility criteria, as shown in Table~\ref{tab:aux_ablation}.
Direct auxiliary transfer already provides strong performance, while introducing the utility criterion yields substantial gains, improving the four metrics by about 2.0 percentage points on average.
This demonstrates the importance of filtering auxiliary evidence that does not benefit SID prediction rather than transferring it indiscriminately.
Further incorporating the accessibility criterion brings consistent additional improvements, confirming that auxiliary capabilities should be transferred only when they are both useful and accessible to the student.
% Utility-aware selection improves retrieval performance, while further considering student accessibility achieves the best results, validating the importance of selectively transferring useful and accessible auxiliary evidence.

\begin{table}
\centering
\caption{Ablation of selective auxiliary-evidence utilization mechanism}
\label{tab:aux_ablation}
\resizebox{\linewidth}{!}{
\begin{tabular}{lcc|cc|cc}
\toprule
\multirow{2}{*}{Method}
& \multicolumn{2}{c|}{Criterion}
& \multicolumn{2}{c|}{CLICK}
& \multicolumn{2}{c}{PURCHASE} \\
& Utility & Accessibility
& H@10 & R@10
& H@10 & R@10 \\
\midrule
Direct Auxiliary Transfer
&  &
& 77.04 & 53.05
& 57.84 & 57.08 \\

+ Utility
& \checkmark &
& 78.71 & 54.91
& 59.70 & 59.06 \\

+ Accessibility
& \checkmark & \checkmark
& \textbf{79.19} & \textbf{55.27}
& \textbf{60.12} & \textbf{59.48} \\
\bottomrule
\end{tabular}}
\end{table}

\section{Conclusion}
In this paper, we have presented \textbf{PailitaoGR} for generative image retrieval, bringing \emph{Latent Think-with-Images} into full-image retrieval by internalizing target-focused perception and selective auxiliary-evidence utilization.
The proposed method consists of two components: target-focused perception mechanism and selective auxiliary-evidence utilization mechanism.
% The proposed target-focused perception mechanism can strengthen search-target understanding by assigning higher importance to visual tokens of the search target and progressively guiding SID decoding toward target-related regions during generation. By internalizing the target-focused perception capability learned from cropped views, it enables the model to identify and attend to the search target directly from the original query image, achieving \textit{Zooming without Cropping}.
% The proposed selective auxiliary-evidence utilization mechanism selectively enhances auxiliary cues that are relevant to the search target and beneficial for SID prediction. By transferring only useful and accessible auxiliary-evidence utilization capability, it enables the model to exploit textual cues from the original image without explicit OCR input, achieving \textit{Reading without OCR}.
The proposed target-focused perception mechanism can strengthen search-target perception by assigning higher importance to target-related visual tokens and progressively guiding SID generation toward discriminative target regions, enabling \textit{Zooming without Cropping}.
The proposed selective auxiliary-evidence utilization mechanism can selectively enhance auxiliary cues that are relevant to the search target and beneficial to SID prediction, enabling \textit{Reading without OCR}.
We further constructed training and evaluation data from real-world online image-search logs.
Extensive experiments show that PailitaoGR consistently outperforms existing retrieval methods and even surpasses both the Crop Teacher and OCR Teacher using only the original query image, demonstrating the effectiveness of the proposed \emph{Latent Think-with-Images} paradigm for generative image retrieval.
% Extensive experiments validate the effectiveness of the proposed \emph{Latent Think-with-Images} paradigm for generative image retrieval.

\newpage
\section{Ethical Considerations}
The industrial dataset contains anonymized user behavior logs, search queries, and item images. These data are used only for the stated research and production purposes, without attempting to identify individual users. Data handling follows platform-defined policies for retention, access control, and deletion. Together, anonymization, purpose limitation, and platform-level governance provide safeguards for responsible model training and evaluation.

% Do not include acknowledgments in the anonymous review submission.
% \begin{acks}
% Acknowledgments belong in the camera-ready version only.
% \end{acks}

\bibliographystyle{ACM-Reference-Format}
\bibliography{references}

\end{document}